\documentclass[11pt,a4paper]{article}
\usepackage[hyperref]{emnlp-ijcnlp-2019}
\usepackage{times}
\usepackage{latexsym}
\usepackage{graphicx, wrapfig}
\usepackage{url}

\usepackage{algorithm}
\usepackage{algorithmic}

\usepackage{bm}
\usepackage{color}
\usepackage{multirow}
\usepackage[font=normalsize,position=bottom]{subfig}
\usepackage{arydshln}
\usepackage{balance}
\usepackage{amsmath}
\usepackage{amssymb}

\aclfinalcopy 

\newcommand\confname{EMNLP-IJCNLP 2019}

\title{Instructions for \confname{} Proceedings}

\author{Zi-Yi Dou, Keyi Yu, Antonios Anastasopoulos \\
  Language Technologies Institute, Carnegie Mellon University \\
  {\tt \{zdou, keyiy, aanastas\}@cs.cmu.edu}}

\date{}
\begin{document}

\title{Investigating Meta-Learning Algorithms for Low-Resource Natural Language Understanding Tasks}


\maketitle

\begin{abstract}
Learning general representations of text is a fundamental problem for many natural language understanding (NLU) tasks. Previously, researchers have proposed to use language model pre-training and multi-task learning to learn robust representations. However, these methods can achieve sub-optimal performance in low-resource scenarios. Inspired by the recent success of optimization-based meta-learning algorithms, in this paper, we explore the model-agnostic meta-learning algorithm (MAML) and its variants for low-resource NLU tasks. We validate our methods on the GLUE benchmark and show that our proposed models can outperform several strong baselines. We further empirically demonstrate that the learned representations can be adapted to new tasks efficiently and effectively.
\end{abstract}

\section{Introduction}

 With the ability to learn rich distributed representations of data in an end-to-end fashion, deep neural networks have achieved the state of the arts in a variety of fields \cite{he2017mask,Vaswani:2017:NIPS,povey2018semi,yu2018deep}. 
 For natural language understanding (NLU) tasks, robust and flexible language representations can be adapted to new tasks or domains efficiently. 
 Aiming at learning representations that are not exclusively tailored to any specific tasks or domains, researchers have proposed several ways to learn general language representations.  

Recently, there is a trend of learning universal language representations via language model pre-training~\cite{dai2015semi,peters2018deep,radford2018improving}. In particular, \newcite{devlin2018bert} present the BERT model which is based on a bidirectional Transformer ~\cite{Vaswani:2017:NIPS}. BERT is pre-trained with both masked language model and next sentence prediction objectives and exhibits strong performance on several benchmarks, attracting huge attention from researchers. Another line of research tries to apply multi-task learning to representation learning~\cite{liu2015representation,luong2015multi}. Multi-task learning allows the model to leverage supervision signals from related tasks and prevents the model from overfitting to a single task. By combining the strengths of both language model pre-training and multi-task learning, ~\newcite{liu2019multi} improve the BERT model with multi-task learning and their proposed MT-DNN model successfully achieves state-of-the-art results on several NLU tasks. 

\begin{figure}[t]
\centering
\includegraphics[width=0.49\textwidth]{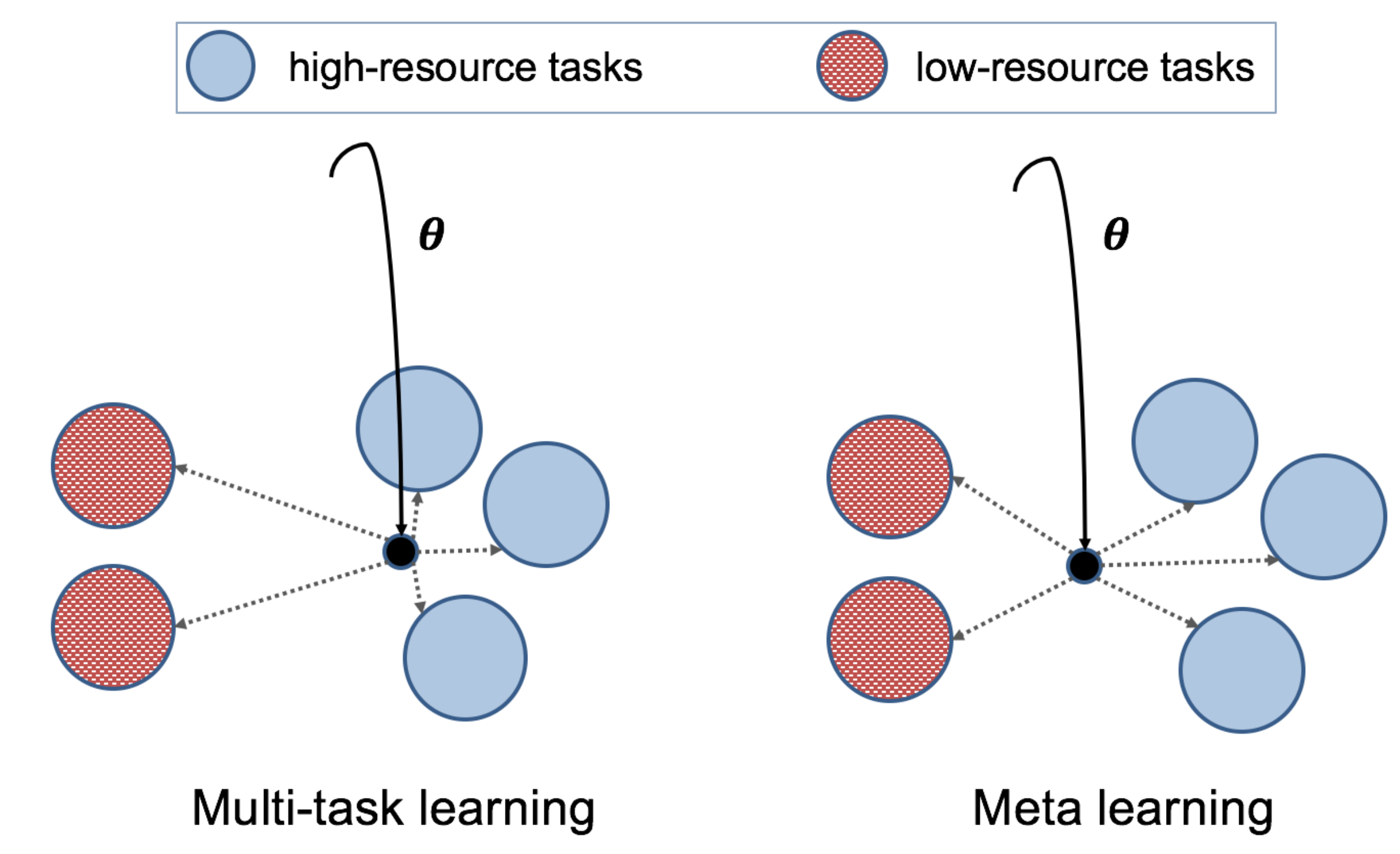}
\caption{\label{fig:1}Differences between multi-task learning and meta learning. Multi-task learning may favor high-resource tasks over low-resource ones while meta-learning aims at learning a good initialization that can be adapted to any task with minimal training samples. The figure is adapted from~\newcite{gu2018meta}.} 
\label{fig:vis}
\end{figure}

Although multi-task learning can achieve promising performance, there still exist some potential problems. As shown in Figure~\ref{fig:1}, multi-task learning may favor tasks with significantly larger amounts of data than others. ~\newcite{liu2019multi} alleviate this problem by adding an additional fine-tuning stage after multi-task learning. In this paper, we propose to apply meta-learning algorithms in general language representations learning. Meta-learning algorithms aim at learning good initializations that can be useful for fine-tuning on various tasks with minimal training data, which makes them appealing alternatives to multi-task learning. Specifically, we investigate the recently proposed model-agnostic meta-learning algorithm (MAML) ~\cite{finn2017model} and its variants, namely first-order MAML and Reptile~\cite{nichol2018first}, for NLU tasks.

We evaluate the effectiveness and generalization ability of the proposed approaches on the General Language Understanding Evaluation (GLUE) benchmark~\cite{wang2018glue}. Experimental results demonstrate that our approaches successfully outperform strong baseline models on the four low-resource tasks. In addition, we test generalization capacity of the models by fine-tuning them on a new task, and the results reveal that the representations learned by our models can be adapted to new tasks more effectively compared with baseline models.

\section{Proposed Approaches}
In this section, we first briefly introduce some key ideas of meta learning, and then illustrate how we apply meta-learning algorithms in language representations learning.

\subsection{Background: Meta Learning}
Meta-learning, or learning-to-learn, has recently attracted researchers' interests in the machine learning community~\cite{lake2015human}. The goal of meta-learning algorithms is to allow fast adaptation on new training data. 
In this paper, we mainly focus on optimization-based meta-learning algorithms, which achieve the goal by adjusting the optimization algorithm. Specifically, we investigate MAML, one of the most representative algorithms in this category, and its variants for NLU tasks. 

MAML and its variants offer a way to learn from a distribution of tasks and adapt to target tasks using few samples.  Formally, given a set of tasks $\{T_1, \cdots, T_k\}$, the process of learning model parameters $\theta$ can be understood as~\cite{gu2018meta}:
$$
\theta^*_t = \text{Learn}(T_t; \text{MetaLearn}(T_1, \cdots, T_k) ),
$$
where $T_t$ is the target task.

Hopefully, by exposing models to a variety of tasks, the models can learn new tasks with
few steps and minimal amounts of data.

\subsection{General Framework}
In this part, we introduce the general framework of the MAML approach and its variants, including first-order MAML and Reptile. 

\begin{algorithm}
\caption{\label{alg:main} Training procedure.}
\begin{algorithmic}
    \STATE{Pre-train model parameters $\theta$ with unlabeled datasets.}
    \WHILE{not done}
        \STATE{Sample batch of tasks $\{T_i\} \sim p(T)$}
        \FORALL{$T_i$}
            \STATE{Compute $\theta^{(k)}_i$ with Eqn.~\ref{eqn:1}.}
        \ENDFOR
        \STATE{Update $\theta$ with Eqn.~\ref{eqn:2}.}
	\ENDWHILE
	\STATE{Fine-tune $\theta$ on the target task.}
\end{algorithmic}
\end{algorithm}

We first describe the meta-learning stage. Suppose we are given a model $f_\theta$ with parameters $\theta$ and a task distribution $p(T)$ over a set of tasks $\{T_1, T_2, \cdots, T_k\}$, at each step during the meta-learning stage, we first sample a batch of tasks $\{T_i\} \sim p(T)$, and then update the model parameters by $k$ $(k \geq 1 )$ gradient descent steps for each task $T_i$ according to the equation:
\begin{equation}
\label{eqn:1}
     \theta^{(k)}_i = \theta^{(k-1)}_i - \alpha \nabla_{\theta^{(k-1)}_i} L_i (f_{\theta_i^{(k-1)}}),
\end{equation}
where $L_i$ is the loss function for $T_i$ and $\alpha$ is a hyper-parameter. 

The model parameters $\theta$ are then updated by:
\begin{equation}
\label{eqn:2}
    \theta = \text{MetaUpdate}(\theta; \{\theta_i^{(k)}\}).
\end{equation}
We would illustrate the {\it MetaUpdate} step in the following part. It should be noted that the data used for the { MetaUpdate} step (Eqn. ~\ref{eqn:2}) is different from that used for the first $k$ gradient descent steps (Eqn. ~\ref{eqn:1}). 

The overall training procedure is shown in Algorithm~\ref{alg:main}. Basically, the algorithm consists of three stages: the pre-training stage as in BERT, the meta-learning stage and the fine-tuning stage.

\subsection{The MetaUpdate Step}
As demonstrated in the previous paragraph, MetaUpdate is an important step in the meta-learning stage. In this paper, we investigate three ways to perform MetaUpdate as described in the following parts.

\paragraph{MAML} The vanilla MAML algorithm~\cite{finn2017model} updates the model with the meta-objective function:
$$
 \min_\theta \sum_{T_i \sim p(T)} L_i(f_{\theta^{(k)}_i})
$$

Therefore, MAML would implement the MetaUpdate step by updating $\theta$ according to:

\begin{equation*}
    \theta = \theta - \beta \sum_{T_i \sim p(T)} \nabla_\theta L_i(f_{\theta^{(k)}_i}),
\end{equation*}
where $\beta$ is a hyper-parameter.

\paragraph{First-Order MAML}  Suppose $\theta^{(k)}$ is obtained by performing $k$ inner gradient steps starting from the initial parameter $\theta^{(0)}$, we can deduce that:
$$
\small
\begin{aligned}
    \nabla_{\theta^{(0)}} L(f_{\theta^{(k)}}) &= \nabla_{\theta^{(k)}} L(f_{\theta^{(k)}}) \prod_{i=1}^k \nabla_{\theta^{(i-1)}} \theta^{(i)} \\
    = \nabla_{\theta^{(k)}}&L(f_{\theta^{(k)}}) \prod_{i=1}^k (I- \alpha \nabla_{\theta^{(i-1)}}^2 L(f_{\theta^{(i-1)}})).  \\
\end{aligned}
$$

Therefore, MAML requires calculating second derivatives, which can be both computationally and memory intensive. First-Order MAML (FOMAML) ignores the second derivative part and implement the MetaUpdate as:
\begin{equation*}
    \theta = \theta - \beta \sum_{T_i \sim p(T)} \nabla_{\theta^{(k)}_i} L_i(\theta^{(k)}_i).
\end{equation*}

\paragraph{Reptile} Reptile~\cite{nichol2018first} is another first-order gradient-based meta-learning algorithm that is similar to joint training, as it implements the MetaUpdate step as:
\begin{equation*}
    \theta = \theta + \beta \frac{1}{|\{T_i\}|} \sum_{T_i \sim p(T)} (\theta^{(k)}_i - \theta).
\end{equation*}

Basically, Reptile moves the model weights towards new parameters obtained by multiple gradient descent steps. Despite the simplicity of Reptile, it has been demonstrated to achieve competitive or superior performance compared to MAML.

\subsection{Choosing the Task Distributions}
We experiment with three different choices of the task distribution $p(T)$. Specifically, we propose the following options: 
\begin{itemize}
    \item {\bf Uniform}: sample tasks uniformly.
    \item {\bf Probability Proportional to Size (PPS)}: the probability of selecting a task is proportional to the size of its dataset.
    \item {\bf Mixed}: at each epoch, we first sample tasks uniformly and then exclusively select the target task. 
\end{itemize}

\begin{table}[t]
  \centering
  \resizebox{1.0\columnwidth}{!}{
  \begin{tabular}{c||c|c|c|c}
  \multirow{2}{*}{\bf Model} & \multicolumn{4}{c}{\bf Test Dataset} \\ 
  & CoLA & MRPC & STS-B & RTE\\
  \hline
  \hline
  BERT & 52.1 & 88.9/84.8 & 87.1/85.8 & 66.4\\
   MT-DNN & 51.7 & 89.9/86.3 & 87.6/86.8 & 75.4 \\
   \hdashline
 MAML & \bf  53.4 &  89.5/85.8 & 88.0/87.3 & 76.4\\
 FOMAML & 51.6 & 89.9/86.4 &88.6/88.0 & 74.1\\ 
 Reptile & 53.2 & \bf 90.2/86.7 & \bf 88.7/88.1 & \bf  77.0\\ 
  \end{tabular}
  }
  \caption{Results on GLUE test sets. Metrics differ per task (explained in Appendix A) but the best result is \textbf{highlighted}.}
  \label{tab:main}
\end{table}

\section{Experiments}
We conduct experiments on the GLUE dataset~\cite{wang2018glue} and only on English. Following previous work~\cite{devlin2018bert,liu2019multi} we do not train or test models on the WNLI dataset~\cite{levesque2012winograd}. We treat the four high-resource tasks, namely {SST-2}~\cite{socher2013recursive}, {QQP},\footnote{data.quora.com/First-Quora-DatasetRelease-Question-Pairs} {MNLI}~\cite{williams2018broad}, and {QNLI}~\cite{rajpurkar2016squad}, as auxiliary tasks. 
The other four tasks, namely {CoLA}~\cite{warstadt2018neural}, {MRPC}~\cite{dolan2005automatically}, {STS-B}~\cite{cerasemeval}, and {RTE~\cite{dagan2005pascal}} are our target tasks. 
We also evaluate the generalization ability of our approaches on the SciTail dataset~\cite{khot2018scitail}. The details of all datasets are illustrated in Appendix A. 

We compare our models with two strong baselines:  the BERT model ~\cite{devlin2018bert} and the MT-DNN model ~\cite{liu2019multi}. While the former pre-trains the Transformer model on large amounts of unlabeled dataset, the latter further improves it with multi-task learning. 

For BERT and MT-DNN, we use their publicly available code to obtain the final results. The setting of MT-DNN is slightly different from the setting of BERT in terms of optimizer choices. We implement our algorithms upon the {\bf BERT}$_\text{\bf BASE}$ model.\footnote{{\bf BERT}$_\text{\bf BASE}$ and {\bf BERT}$_\text{\bf LARGE}$ differ at the number of hidden layers (12 vs. 24), hidden size (768 vs. 1024) and the number of attention heads (12 vs. 16).}  
We use the Adam optimizer~\cite{kingma2015adam} with a batch size of 32 and learning rates of 5e-5 to train the models for 5 epochs in the meta-learning stage. We set the update step $k$ to 5, the number of sampled tasks in each step to 8 and $\alpha$ to 1e-3. 


\subsection{Results}

We first use the three meta-learning algorithms with PPS sampling and present in Table~\ref{tab:main} the experimental results on the GLUE test set. Generally, the meta-learning algorithms achieve better performance than the strong baseline models, with Reptile performing the best. 

Since the MT-DNN also uses PPS sampling, the improvements suggest meta-learning algorithms can indeed learn better representations compared with multi-task learning. Reptile outperforming MAML indicates that reptile is a more effective and efficient algorithm compared with MAML in our setting.


\subsection{Ablation Studies}
\paragraph{Effect of Task Distributions}
\begin{table}[t]
  \centering
  \resizebox{1.0\columnwidth}{!}{
  \begin{tabular}{c||c|c|c|c}
  \bf Model & CoLA & MRPC & STS-B & RTE\\
  \hline
  \hline
 Reptile-PPS & \bf 61.6 & \bf  90.0 & \bf  90.3 & \bf  83.0  \\
 Reptile-Uniform &61.5 & 84.0 & \bf 90.3 & 75.7\\ 
 Reptile-Mixed 2:1 & 60.3 &  87.8 & \bf  90.3 & 71.0   \\ 
  Reptile-Mixed 5:1 & \bf 61.6 &  85.8 & 90.1 & 74.7   \\ 
  \end{tabular}
  }
  \caption{Effect of task distributions. We report the accuracy or Matthews correlation on development sets.}
  \label{tab:dis}
\end{table}

As we have mentioned above, we propose three different choices of the task distribution $p(T)$ in this paper. Here we train Reptile with these task distributions and test models' performance on the development set as shown in Table~\ref{tab:dis}. 

For uniform sampling, we set the number of training steps equal to that of the PPS method. For mixed sampling, we try mix ratios of both 2:1 and 5:1. 
The results demonstrate that Reptile with PPS sampling achieves the best performance, which suggests that larger amounts of auxiliary task data can generally lead to better performance.

\begin{table}[t]
  \centering
  \resizebox{1.0\columnwidth}{!}{
  \begin{tabular}{c||c|c|c|c|c|c}
  \bf Model & \#Upt & $\alpha$ & CoLA & MRPC & STS-B & RTE\\
  \hline
  \hline
  \multirow{5}{*}{Reptile}   & 3 &  \multirow{1}{*}{1e-3} & 60.7 & 89.7 & 90.2 & 77.9 \\
    \cline{2-7}
& \multirow{3}{*}{5} &   1e-4   & \bf 62.0 & 88.0 & 90.1 & 81.2   \\
&   & 1e-3 &  61.6 & \bf  90.0 &  \bf 90.3 & \bf  83.0  \\
&   & 1e-2 & 60.1 & 87.8 & 89.5 & 73.9 \\
  \cline{2-7}
    & 7 & 1e-3  & 57.8 & 88.7 & 90.0 & 81.4  \\ 
  \end{tabular}
  }
  \caption{Effect of the number of update steps and the inner learning rate $\alpha$.}
  \label{tab:ab2}
\end{table}

\paragraph{Effect of Hyperparameters for Meta-Gradients}
In this part, we test the effect of the number of update steps $k$ and the learning rate in the inner learning loop. The experimental results on the development sets are shown in Table~\ref{tab:ab2}. We find that setting $k$ to 5 is the optimal strategy and more or fewer update steps may lead to worse performance. 

Smaller $k$ would make the algorithms similar to joint training as joint training is an extreme case of Reptile where $k=1$, and thus cause the model to lose the advantage of using meta-learning algorithms. Similarly, Larger $k$ can make the resulting gradients deviate from the normal ones and become uninformative.

We also vary the inner learning rate $\alpha$ and investigate its impact. The results are listed in Table~\ref{tab:ab2}. We can see that larger $\alpha$ may degrade the performance because the resulting gradients deviate a lot from normal ones. 
The above two ablations studies demonstrate the importance of making the meta-gradient informative. 

\subsection{Transferring to New Tasks}
\begin{figure}[t]
\centering
\includegraphics[width=0.45\textwidth]{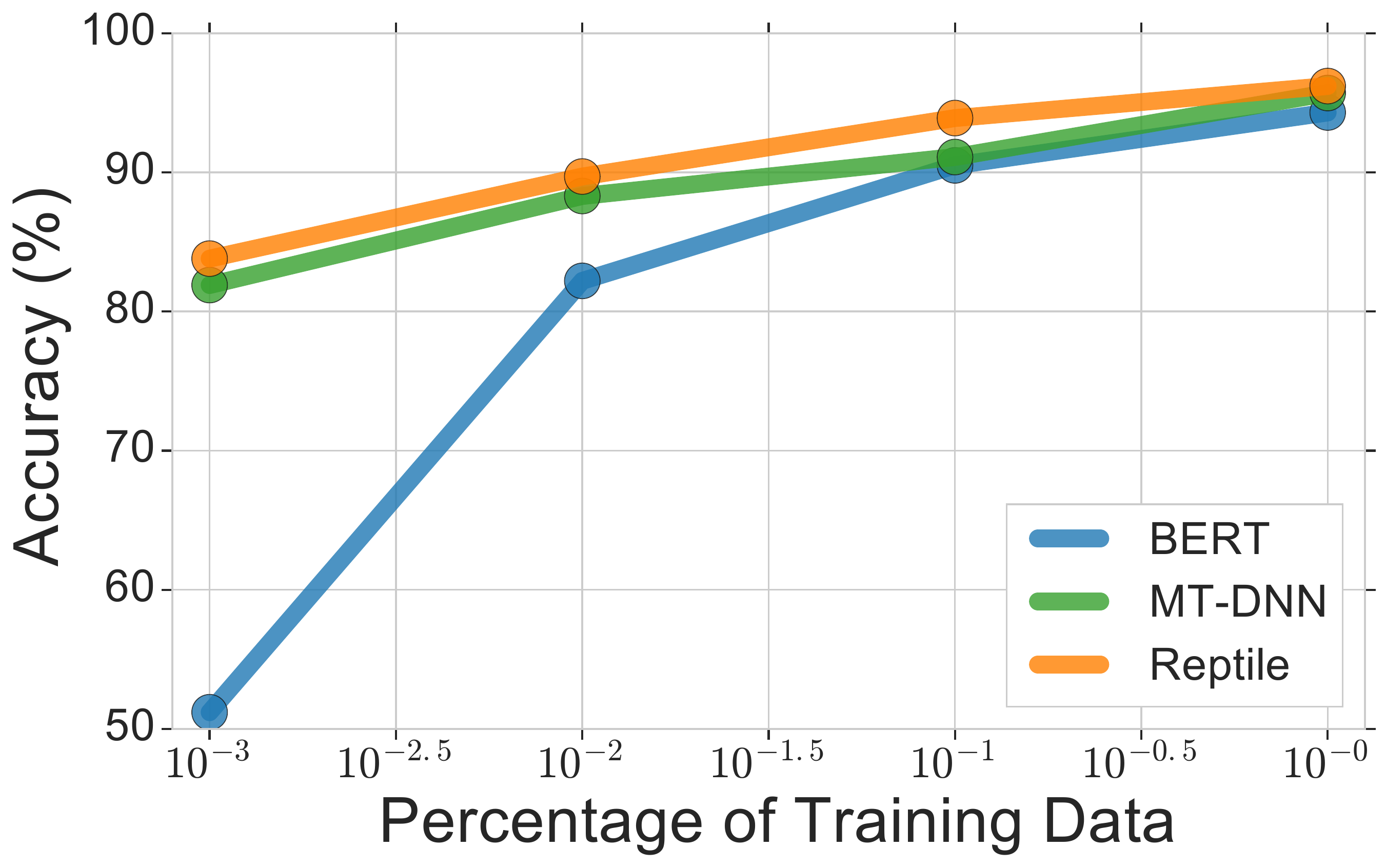}
\caption{\label{fig:transfer} Results on transfer learning. The target task is SciTail which the model does not come across during the meta-learning stage.}
\end{figure}

In this part, we test whether our learned representations can be adapted to new tasks efficiently. To this end, we perform transfer learning experiments on a new natural language inference dataset, namely SciTail. 

We randomly sample 0.1\%, 1\%, 10\% and 100\% of the training data and test models' performance on these datasets. Figure~\ref{fig:transfer} reveals that our model consistently outperforms the strong {MT-DNN} baseline across different settings, indicating the learned representations are more effective for transfer learning. In particular, the algorithm is more effective when less data are available, especially compared to BERT, suggesting the meta-learning algorithms can indeed be helpful for low-resource tasks. 

\section{Related Work}
There is a long history of learning general language representations. 
Previous work on learning general language representations focus on learning word~\cite{mikolov2013distributed,pennington2014glove} or sentence representations~\cite{le2014distributed,kiros2015skip} that are helpful for downstream tasks. 
Recently, there is a trend of learning contextualized word embeddings~\cite{dai2015semi,mccann2017learned,peters2018deep,howard2018universal}. One representative approach is the BERT model~\cite{devlin2018bert} which learns contextualized word embeddings via bidirectional Transformer models.

Another line of research on learning representations focus on multi-task learning~\cite{collobert2011natural,liu2015representation}. In particular, ~\newcite{liu2019multi} propose to combine multi-task learning with language model pre-training and demonstrate the two methods are complementary to each other.

Meta-learning algorithms have received lots of attention recently due to their effectiveness~\cite{finn2017model,fan2018learning}. However, the potential of applying meta-learning algorithms in NLU tasks have not been fully investigated yet. ~\newcite{gu2018meta} have tried to apply first-order MAML in machine translation and ~\newcite{qian2019domain} propose to address the domain adaptation problem in dialogue generation by using MAML. To the best of our knowledge, the Reptile algorithm, which is simpler than MAML and potentially more useful, has been given less attention.

\section{Conclusion}
In this paper, we investigate three optimization-based meta-learning algorithms for low-resource NLU tasks. We demonstrate the effectiveness of these algorithms and perform a fair amount of ablation studies. We also show the learned representations can be adapted to new tasks effectively. Our study suggests promising applications of meta-learning algorithms in the field of NLU. 
Future directions include integrating more sophisticated training strategies of meta-learning algorithms as well as validating our algorithms on other datasets.

\section*{Acknowledgements}
The authors are grateful to the anonymous reviewers for their constructive comments, and to Graham Neubig and Junxian He for helpful discussions.
This material is based upon work generously supported partly by the National Science Foundation under grant 1761548.

\bibliographystyle{acl_natbib}
\bibliography{emnlp-ijcnlp-2019.bib}

\appendix

\begin{table*}[t]
  \centering
  \resizebox{1.7\columnwidth}{!}{
  \begin{tabular}{c||c|c|c|c}
\bf Corpus & \bf Task & \bf \# Train & \bf \# Label & \bf Metrics \\
 \hline
 \hline
 \multicolumn{5}{c}{\it Single-Sentence Tasks} \\
 \hline
 CoLA & Acceptability & 8.5k & 2 & Matthews correlation \\
 \hdashline
 SST-2 & Sentiment & 67k & 2 & Accuracy \\
 \hline
 \hline
 \multicolumn{5}{c}{\it Similarity and Paraphrase Tasks} \\
 \hline
 MRPC & Paraphrase & 3.7k & 2 & F1/Accuracy \\
 \hdashline
 STS-B & Similarity & 7k & 1 & Pearson/Spearman correlation \\
 \hdashline
  QQP & Paraphrase & 364k & 2 & F1/Accuracy \\
  \hline
 \hline
 \multicolumn{5}{c}{\it Inference Tasks} \\
 \hline
 MNLI & NLI & 393k & 3 & Accuracy \\
 \hdashline
 QNLI & QA/NLI & 105k & 2 & Accuracy \\
 \hdashline
  RTE & NLI &2.5k & 2 & Accuracy \\
  \hdashline
   WNLI & NLI &634 & 2 & Accuracy \\
 \hdashline
  SciTail & NLI & 23.5k & 2 & Accuracy \\
 \hline
  \end{tabular}
  }
  \caption{Basic information and statistics of the GLUE and SciTail datasets~\cite{williams2018broad}.}
  \label{tab:glue}
\end{table*}

\section{The GLUE and SciTail Datasets}
\label{apd:1}

Basically, the GLUE dataset ~\cite{wang2018glue} consists of three types of tasks: single-sentence classification, similarity and paraphrase tasks, and inference tasks, as shown in Table~\ref{tab:glue}. 

\paragraph{Single-Sentence Classification.} The model needs to make a prediction given a single sentence for this type of tasks. The goal of the {\bf CoLA} task is to predict whether an English sentence is grammatically plausible and the goal of the {\bf SST-2} task is to determine whether the sentiment of a sentence is positive or negative. 

\paragraph{Similarity and Paraphrase Tasks.} For this type of tasks, the model needs to determine whether or to what extent two given sentences are semantically similar to each other. Both the {\bf MRPC} and the {\bf QQP} tasks are classification tasks that require the model to predict whether the sentences in a pair are semantically equivalent. The {\bf STS-B} task, on the other hand, is a regression task and requires the model to output a real-value score representing the semantic similarity of the two sentences.

\paragraph{Inference Tasks.} Both the {\bf RTE} and the {\bf MNLI} tasks aim at predicting whether a sentence is entailment, contradiction or neutral with respect to the other. {\bf QNLI} is converted from a question answering dataset, and the task is to determine whether the context sentence contains the answer to the question. {\bf WNLI}  is to predict if the sentence with the pronoun substituted is entailed by the original sentence. Because the test set is imbalanced and the development set is adversarial, so far none of the proposed models could surpass the performance of the simple majority voting strategy. Therefore, we do not use the WNLI dataset in this paper. 

{\bf SciTail} is a textual entailment dataset that is derived from a science question answering dataset~\cite{khot2018scitail}. Given a premise and a hypothesis, the model need to determine whether the premise entails the hypothesis. The dataset is fairly difficult as the sentences are linguistically challenging and the lexical similarity of premise and hypothesis is high.

\section{Implementation Details}
\label{apd:implement}
Our implementation is based on the PyTorch implementation of BERT.\footnote{https://github.com/huggingface/pytorch-pretrained-BERT}
We first load the pre-trained {\bf BERT}$_\text{\bf BASE}$ model.
We use the Adam optimizer~\cite{kingma2015adam} with a batch size of 32 for both meta-learning and fine-tuning. We set the maximum length to 80 to reduce GPU memory usages. 

In the meta-learning stage, we use a learning rate of 5e-5 to train the models for 5 epochs. Both the dropout and the warm-up ratio are set to 0.1 and we do not use gradient clipping. 
We set the update step $k$ to 5, the number of sampled tasks in each step to 8 and $\alpha$ to 1e-3.

For fine-tuning, again the dropout and warum-up ratio are set to 0.1 and we do not use gradient clipping. The learning rate is selected from \{5e-6, 1e-5, 2e-5, 5e-5\} and the number of epochs is selected from \{3, 5, 10, 20\}. 
We select hyper-parameters that achieve the best performance on the development set. 

We do not use the stochastic answer network as in MT-DNN for efficiency.


\section{Linguistic Information}
\begin{table*}[t]
  \centering
  \resizebox{2.1\columnwidth}{!}{
  \begin{tabular}{c||c|c|c|c|c|c|c|c|c|c}
  \multirow{2}{*}{\bf Model} & \multicolumn{2}{c|}{\bf Surface} & \multicolumn{4}{c|}{\bf Syntactic} & \multicolumn{4}{c}{\bf Semantic} \\
  \cline{2-11}
  & SentLen&Word& TreeDep &ToCo & BShif & Tense & SubNum & ObjNum & SOMO & CoIn \\
  \hline
  \hline
 Majority Voting & 16.67 & 0.10 & 17.88 & 5.00 & 50.00 & 50.00 & 50.00 & 50.00 & 50.13 & 50.00 \\
 \hdashline
 BERT-Layer 1 & 90.84 & 7.54 & 32.31 & 57.91 & 50.67 & 78.83 & 77.50 & 75.65 & 50.13 & 50.05\\
 BERT-Layer 6 & 69.87 & 1.06 & 31.96 & 76.97 & 79.66 & 86.19 & 84.33 & 77.58 & 57.73 & 63.43 \\
 BERT-Layer 12 & 63.15 & 32.98 & 28.80 & 71.36 & 85.67 & 89.72 & 76.63 & 76.52 & 60.92 & 70.91\\
 \hdashline 
 MTDNN-Layer 1 & 92.43 & 25.84 & 33.57 & 58.64 & 50.00 & 78.00 & 80.70 & 79.83 & 51.26 & 51.57\\
 MTDNN-Layer 6 & 80.11 & 21.41 & 31.73 & 59.58 & 76.00 & 81.89 & 80.36 & 80.00 & 55.52 & 58.31\\
 MTDNN-Layer 12 & 58.15 & 23.49 & 28.03 & 56.93 & 75.58 & 85.47 & 76.94 & 72.76 & 58.16 & 66.09\\
  \hdashline 
 MAML-Layer 1 & 92.21 & 2.09 & 30.64 & 55.27 & 50.00 & 77.71 & 72.61 & 70.44 & 50.13 & 52.49 \\
 MAML-Layer 6 & 
 76.26 & 32.13 & 28.24 & 67.45 & 68.43 & 87.88 & 80.79 & 80.07 & 55.40 & 59.38\\
 MAML-Layer 12 & 61.50 & 20.32 & 27.31 & 60.15 & 79.47 & 85.56 & 77.60 & 75.86 & 56.76 & 63.59 \\
  \hdashline 
 FOMAML-Layer 1 & 88.39 & 3.22 & 30.91 & 51.01 & 49.97 & 79.56 & 74.53 & 71.28 & 50.13 & 50.00  \\
 FOMAML-Layer 6 & 
 81.33 & 22.63 & 30.44 & 69.48 & 77.01 & 88.89 & 81.81 & 80.18 &  57.93 & 60.11\\
 FOMAML-Layer 12 & 62.93 & 30.84 & 28.33 & 59.15 & 79.96 & 87.60 & 79.33 & 77.98 & 58.05 & 64.58 \\
  \hdashline 
 Reptile-Layer 1 & 87.97 & 3.26 & 30.00 & 52.88 & 50.74 & 80.48 & 74.32 & 70.90 & 50.13 & 50.00 \\
 Reptile-Layer 6 & 77.55 & 24.52 & 30.74 & 69.18 & 75.20 & 88.42 & 82.11 & 81.03 & 58.52 & 61.39\\
 Reptile-Layer 12 & 60.02 & 29.07 & 27.78 & 59.00 & 82.95 & 87.34 & 77.75 & 75.21 & 59.23 & 67.60 \\
  \end{tabular}
  }
  \caption{Accuracy numbers on the 10 probing tasks~\cite{conneau2018you}.}
  \label{tab:ling}
\end{table*}

In this part, we use 10 probing tasks~\cite{conneau2018you}  to study what linguistic information is captured by each layer of the models. 

A probing task is a classification problem that requires the model to make predictions related to certain linguistic properties of sentences. The abbreviations for the 10 tasks are listed in Table~\ref{tab:ling}. Basically, these tasks are set to test the model's abilities to capture surface, syntactic or semantic information. We refer the reader to~\newcite{conneau2018you} for details. We freeze all the parameters of the models and only train the classification layer for the probing tasks.

First, we can see that the BERT model captures more surface, syntactic and semantic information than other models, suggesting it learns more general representations. MT-DNN and our models, on the other hand, learn representations that are more tailored to the GLUE tasks. 

Second, our models perform better than MT-DNN on the probing tasks, indicating meta-learning algorithms may find a balance between general linguistic information and task-specific information. Among the three meta-learning algorithms, Reptile can capture more general linguistic information than others. Considering Reptile has outperformed the other two models on the GLUE dataset, these results further demonstrate Reptile may be more suitable for NLU tasks.

Third, we find that there may not always exist a monotonic trend on what linguistic information each layer captures. Also, contrary to the findings from ~\newcite{liu2019linguistic} which suggest the middle layers of BERT are more transferable and contain more syntactic and semantic information, our experimental results demonstrate that this may not always be true. We conjecture this is because both syntactic and semantic information are broad concepts and the probing tasks in~\newcite{liu2019linguistic} may not cover all of them. For example, there exist a monotonic trend for SOMO while the middle layers of these models are better at tasks like SubNum.

Another interesting thing to note is that the lower layers of models perform rather poorly on the word content task, which tests whether the model can recover information about the original words in the sentence. We attribute this phenomenon to the use of subwords and position/token embeddings. In the higher layers, the model may gain more word-level information through the self-attention mechanism.
\newpage
\cleardoublepage

\end{document}